\documentclass[balance,upint,subscriptcorrection,varvw,mathalfa=cal=boondoxo,spanish,french,vietnamese,russian,greek,pdf-a,colorlinks]{asmeconf}

\hypersetup{%
	pdfauthor={John H. Lienhard},
	pdftitle={ASME Conference Paper LaTeX Template},                  
	pdfkeywords={ASME conference paper, LaTeX template, BibTeX style},
	pdfsubject = {Describes the asmeconf LaTeX template},			  
	pdflicenseurl={https://ctan.org/pkg/asmeconf},
}

\begin{document}


\ConfName{Proceedings of the ASME 2023\linebreak International Design Engineering Technical Conferences and\linebreak Computers and Information in Engineering Conference}
\ConfAcronym{IDETC/CIE2023}
\ConfDate{August 20--23, 2023} 
\ConfCity{Boston, Massachusetts} 
\PaperNo{IDETC/CIE2023-111245}


\title{Mobile Manipulation Platform for Autonomous Indoor Inspections in Low-Clearance Areas} 
 
%
%
%

\SetAuthors{%
	Erik Pearson\affil{1}, 
	Paul Szenher\affil{1}, 
        Christine Huang\affil{1},
	Brendan Englot\affil{1}
	}

\SetAffiliation{1}{Department of Mechanical Engineering, Stevens Institute of Technology, Hoboken, NJ, USA}


\maketitle

\versionfootnote{Documentation for \texttt{asmeconf.cls}: Version~\versionno, \today.}




\begin{abstract}

Mobile manipulators have been used for inspection, maintenance and repair tasks over the years, but there are some key limitations. Stability concerns typically require mobile platforms to be large in order to handle far-reaching manipulators, or for the manipulators to have drastically reduced workspaces to fit onto smaller mobile platforms. Therefore we propose a combination of two widely-used robots, the Clearpath Jackal unmanned ground vehicle and the Kinova Gen3 six degree-of-freedom manipulator. The Jackal has a small footprint and works well in low-clearance indoor environments. Extensive testing of localization, navigation and mapping using LiDAR sensors makes the Jackal a well developed mobile platform suitable for mobile manipulation. The Gen3 has a long reach with reasonable power consumption for manipulation tasks. A wrist camera for RGB-D sensing and a customizable end effector interface makes the Gen3 suitable for a myriad of manipulation tasks. Typically these features would result in an unstable platform, however with a few minor hardware and software modifications, we have produced a stable, high-performance mobile manipulation platform with significant mobility, reach, sensing, and maneuverability for indoor inspection tasks, without degradation of the component robots' individual capabilities. These assertions were investigated with hardware via semi-autonomous navigation to waypoints in a busy indoor environment, and high-precision self-alignment alongside planar structures for intervention tasks.
\end{abstract}


\section{Introduction}

\begin{figure}[ht]
\centering\includegraphics[width=0.5\linewidth]{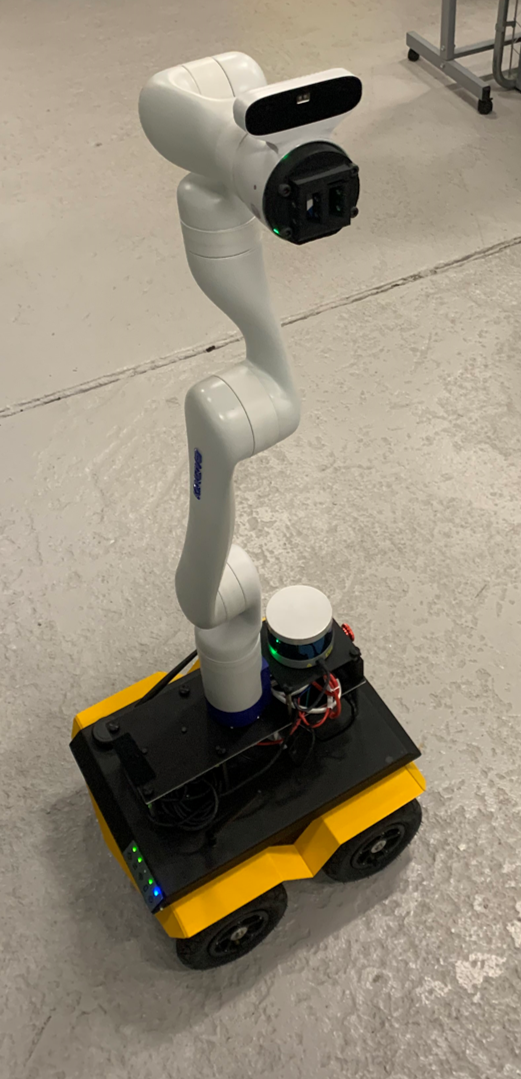}
\caption{Reaching capabilities with small footprint.}
\label{fig:reach}
\end{figure}

Mobile robots have been used for many inspection, maintenance and repair tasks and come in many forms to optimize the tasks being performed. In just the energy sector alone, hundreds of robots have been designed for such tasks \cite{cigre2020application}. One special platform that has the potential to address a multitude of tasks using the same hardware is the mobile manipulation robot. Mobile manipulation requires two main hardware components: a base vehicle that can navigate throughout an environment, and a manipulator that can intervene in the environment. 

Mobile manipulation has been around for more than 30 years \cite{yamamoto1992coordinating} \cite{khatib1999mobile} as an autonomous solution for tasks commonly performed by people. However, many tasks have specific hardware requirements to be accomplished, which has resulted in an abundance of mobile manipulation platforms. Larger mobile manipulators such as the Centauro \cite{kashiri2019centauro}, EL-E \cite{jain2010assistive}, or Care-O-bot \cite{graf2009robotic} are designed for retrieval and carrying tasks, while smaller options such as Stretch \cite{kemp2022design}, Omnivil \cite{engemann2020omnivil}, and youBot-a \cite{bischoff2011kuka} often focus more on interaction with a limited environment. The trend focuses on stability, where larger bases can use larger manipulators, while smaller bases are drastically limited in their manipulation capabilities.

Many robotics teams have built custom mobile manipulation platforms, with varied applications in mind. In this paper, we describe a platform built to have a small footprint for use within a busy indoor environment, while still being able to access a large manipulator workspace. For that purpose, Clearpath's Jackal Unmanned Ground Vehicle (UGV) was chosen for the base. Less than two feet in each dimension, 
this UGV can navigate through busy corridors as easily as a person would. Our Jackal is equipped with a Velodyne VLP-16 ``puck" LiDAR, commonly used for mapping and localization. The Kinova Gen3 6-DOF robotic arm was chosen as the manipulator for its 891mm reach, wrist-mounted RGB-D camera, and customizable wrist interface.

The complete mobile manipulation architecture described in this paper offers an alternative to prior work by incorporating a long reaching manipulator, multiple sensing modalities, and a base that can navigate through cluttered indoor environments effectively.
In the following sections, the hardware and software architectures are described first. Two key performance capabilities 
are then reviewed for compatibility with the platform, including a confirmation that no degradation of individual subsystems occurs. Finally,  recommendations are included for further studies of the platform.


\section{Hardware Architecture}

\begin{figure}[ht]
\centering\includegraphics[width=0.98\linewidth]{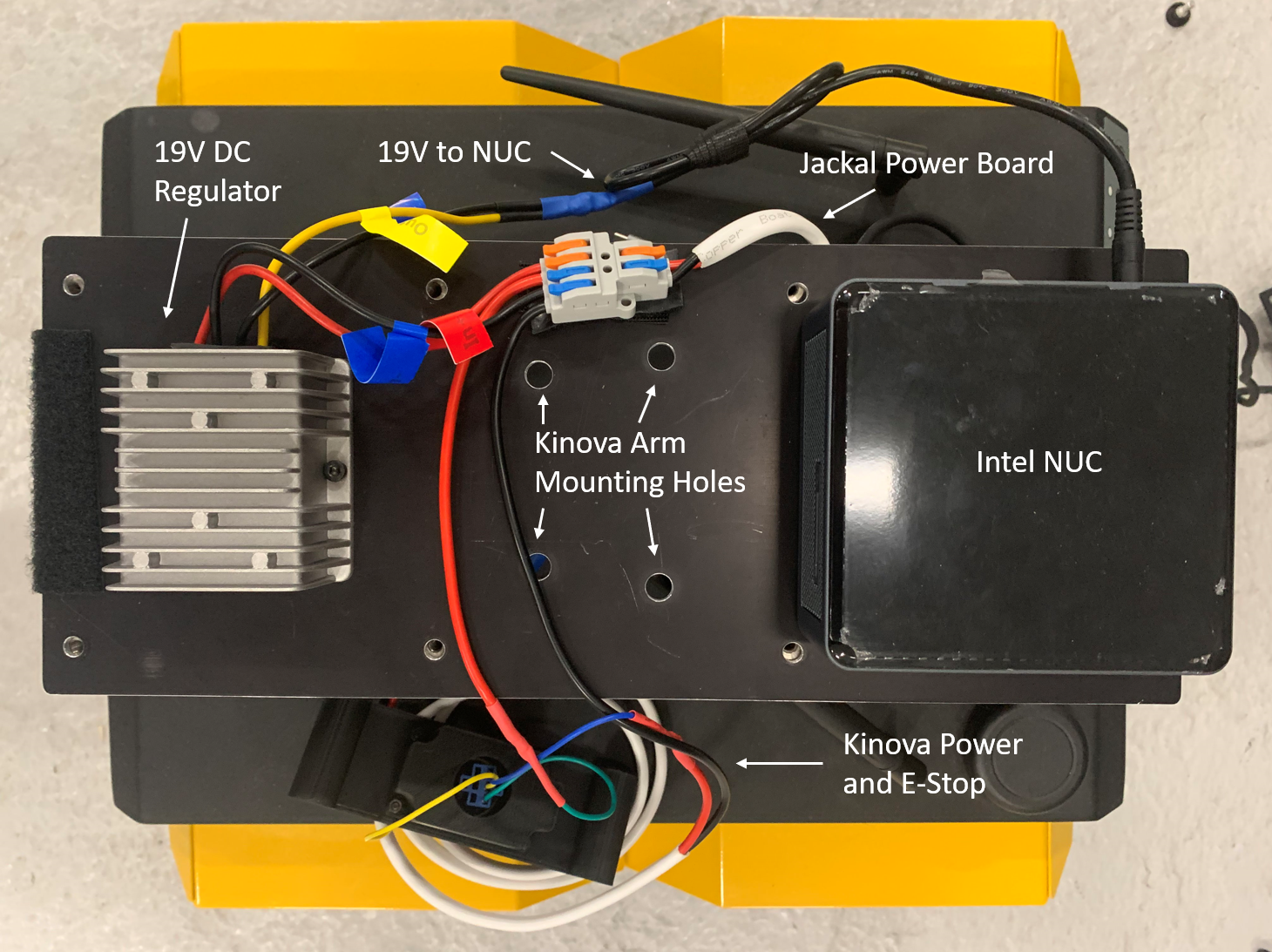}
\caption{Power connections used for mounted hardware. Mounting plate is upside-down for visibility.}
\label{fig:power}
\end{figure}

The Jackal UGV has a surface plate with mounting holes 12cm apart to enable custom hardware integration. For stability, a new plate using all eight mounting holes was fabricated to create a higher platform which would be used as the base of the manipulator. The primary reason for this was to contain all the extra pieces of hardware underneath the plate, while a secondary benefit was to extend the range of the manipulator without compromising stability. Mounted underneath the custom plate on one side is a 19V DC voltage regulator used to power an Intel NUC. The other side has the NUC, which is used to control the manipulator and communicate with the Jackal's onboard computer. Their mounting positions and power connections can be seen in Fig.~\ref{fig:power}, where the plate is flipped for visibility. The batteries provided by Clearpath operate the Jackal from 29.4V to roughly 25.6V. The Kinova Gen3 arm can be powered directly by these voltages, while the NUC requires 19V consistently. When all these electronics are powered simultaneously, a fully charged battery will last approximately 100 minutes.

\begin{figure}[ht]
\centering\includegraphics[width=0.98\linewidth]{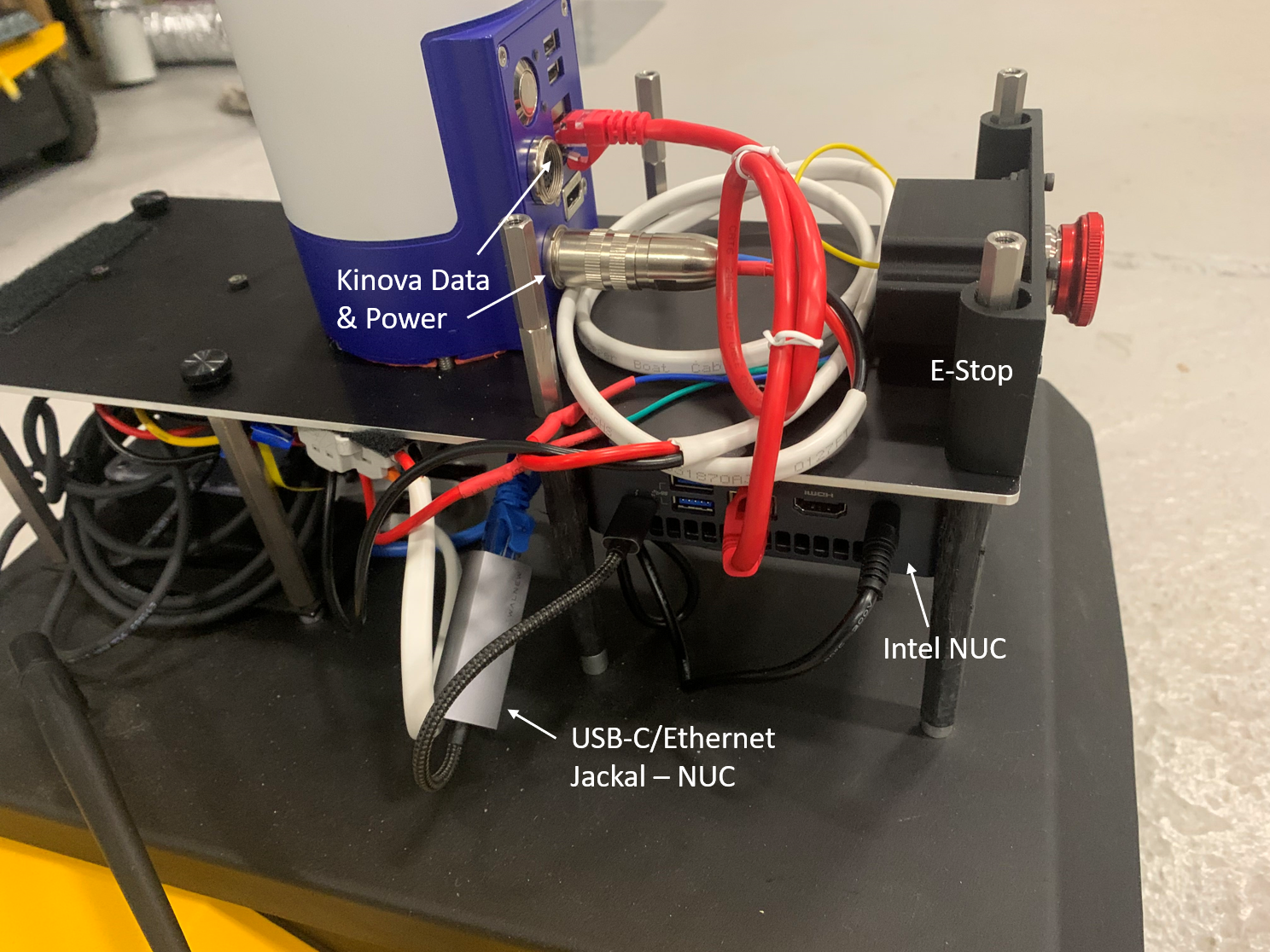}
\caption{Data connections used for mounted hardware.}
\label{fig:data}
\vspace{-4mm}
\end{figure}

Data connections can be seen in Fig.~\ref{fig:data} with the plate and manipulator mounted. The manipulator is centered on the Jackal, which leaves the front or rear for mounting the 16-beam Velodyne puck LiDAR. Obstacle avoidance requires no occlusion of data from the front, therefore the Velodyne was mounted towards the front of the UGV. The NUC only has a single ethernet port, which is used by the Gen3 manipulator, so a USB-C to ethernet dongle was used to connect to the onboard Jackal computer. An E-Stop switch was mounted to the front of the vehicle to cut power to the Gen3 manipulator if necessary.

\begin{figure*}[ht]
\centering\includegraphics[width=0.95\linewidth]{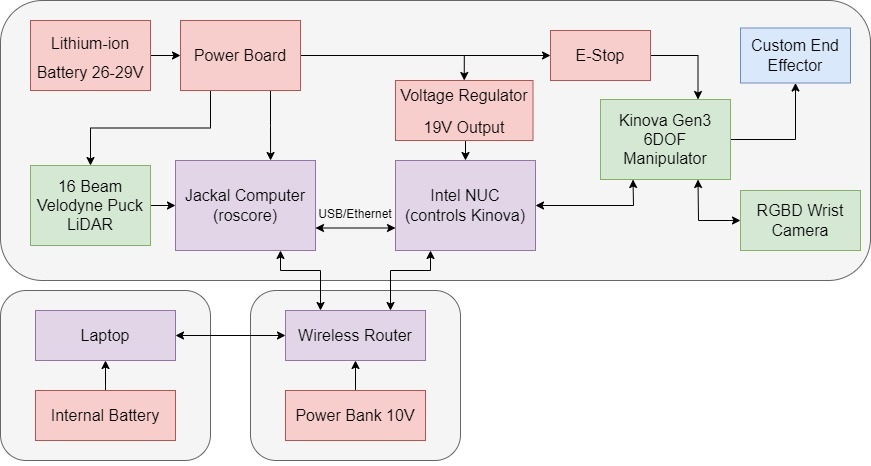}
\caption{Schematic of hardware components. Grey boxes denote separate physical modules, pink describes power supplies, purple are computers, green represents sensors with feedback, while blue defines actuators.}
\label{fig:scheme}
\end{figure*}

A diagram of the complete hardware architecture can be seen in Fig.~\ref{fig:scheme}. While the direct data connection between the Jackal's onboard computer and the NUC ensures no data loss, beginning autonomous exercises requires some form of remote activation. Therefore both computers were connected to a standalone wireless router powered by a battery that can be carried around easily in a backpack when testing. A laptop was configured to connect to the same WiFi to allow terminal control of both the Jackal and the NUC. There were some instances where the Jackal did not connect to the WiFi, however it was possible to access its computer via the NUC, reducing potential failure scenarios.

The integrated mobile manipulation platform possesses three key sensors, seen in green in Fig.~\ref{fig:scheme}. The Velodyne LiDAR is predominantly used for localization and mapping, while the RGB-D camera on the wrist of the Gen3 manipulator handles vision and range sensing tasks relevant to the arm. The manipulator also has torque and position sensors throughout, which enable precise control over its position and effort. The wrist interface can support additional sensors, but is more commonly used for actuation. In this paper, no specific end effector is adopted.

\section{Software Implementation}

Communication between the Jackal onboard computer and the NUC was handled through the Robot Operating System (ROS). All systems have Ubuntu 20.04 installed and use the Noetic distribution of ROS, which enables users to run python3 within ROS. The wired connection on the Jackal has a default static IP that was not changed, while the NUC had to be assigned an IP address for ROS. These wired addresses were different from those used by the wireless router and seen by the laptop. However, with the appropriate configuration, ROS was able to communicate seamlessly between all three devices with the Jackal as the only ROS master. Messages that were sent from any of the three devices were received by all three devices without packet loss until larger data sizes were attempted over WiFi. The chance of losing data over spotty WiFi was the primary reason for including a wired connection between the Jackal's onboard computer and the NUC, ensuring that locally run programs never lose data.

While the Jackal boots ROS upon startup, the NUC and Gen3 manipulator both require manual startup. The NUC itself operates as an ordinary computer with no monitor nor input mechanisms. By pressing the power button, the NUC will boot up and wait. The  manipulator will operate the same way, booting up with the power button before waiting for a command. Accordingly, the laptop is crucial for initiating automation. Using the laptop to secure shell (SSH) into the NUC over WiFi allows users to start the ROS programs installed. The Jackal powerboard only releases power once the Jackal has been turned on, so the NUC can only ever be turned on afterwards. Therefore the ROS program initiated on the NUC to control the Gen3 manipulator will see the Jackal ROS master when attempting to start. Once the Jackal is on and the NUC program is running, the systems are integrated and functional.

\begin{figure}[ht]
\centering\includegraphics[width=0.98\linewidth]{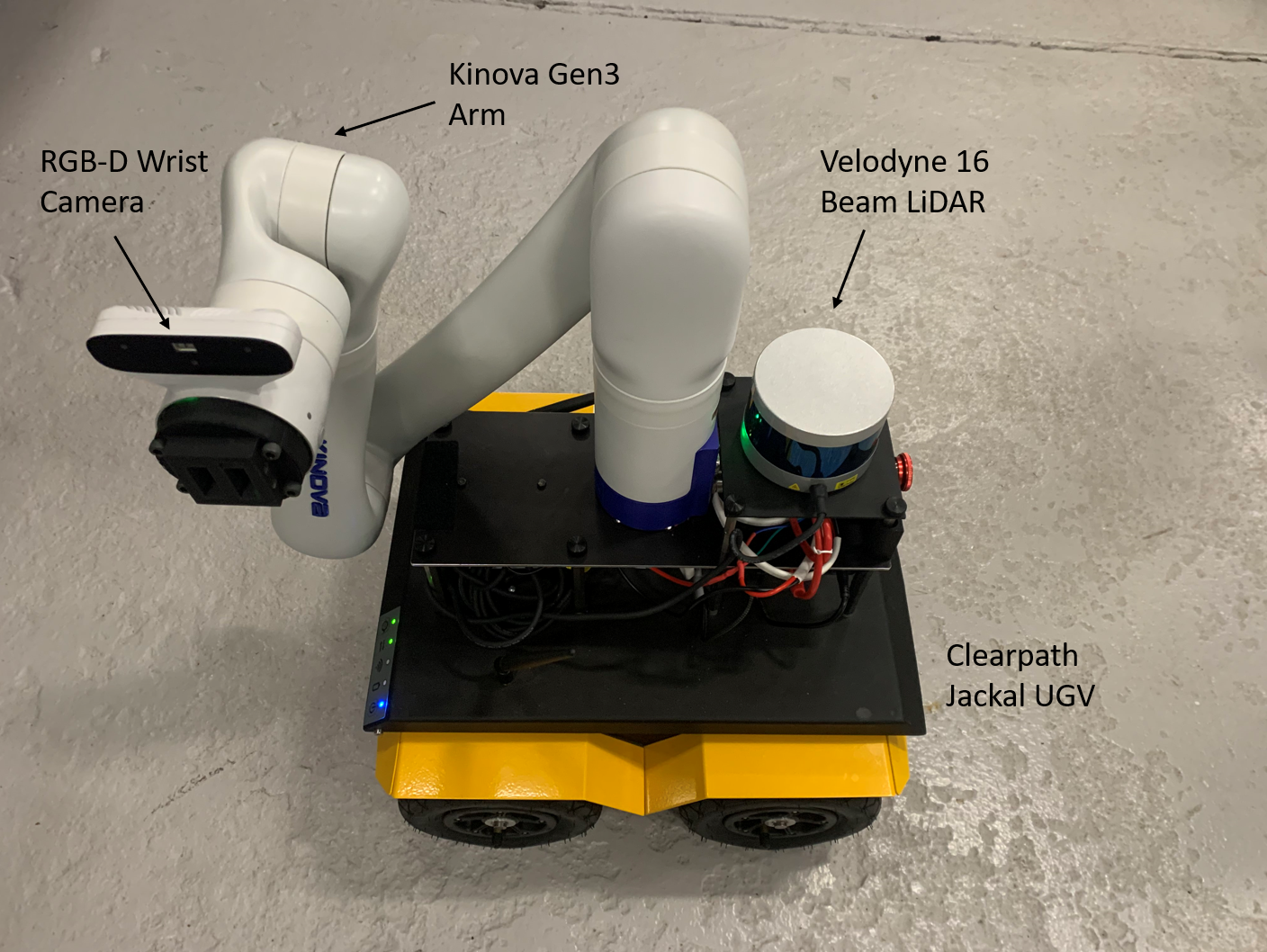}
\caption{Travel configuration for safe, stable driving throughout the environment.}
\label{fig:travel}
\end{figure}

\begin{figure*}[ht]
\begin{subfigure}[t]{0.49\textwidth}
\centering{
  \includegraphics[width=\textwidth]{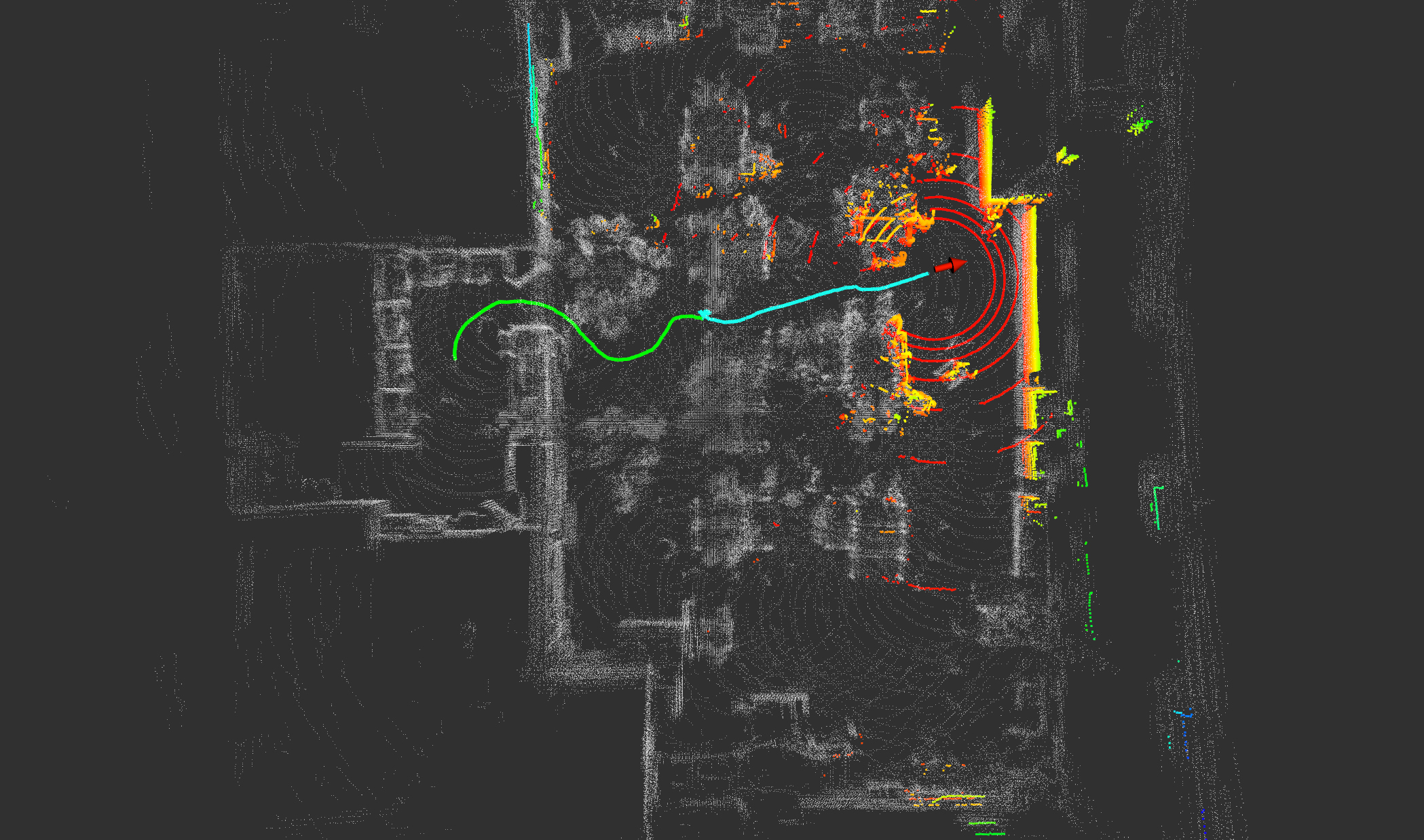}
}%
\subcaption{First semi-autonomous waypoint}
\end{subfigure}%
\hfill
\begin{subfigure}[t]{0.49\textwidth}
\centering{
\includegraphics[width=\textwidth]{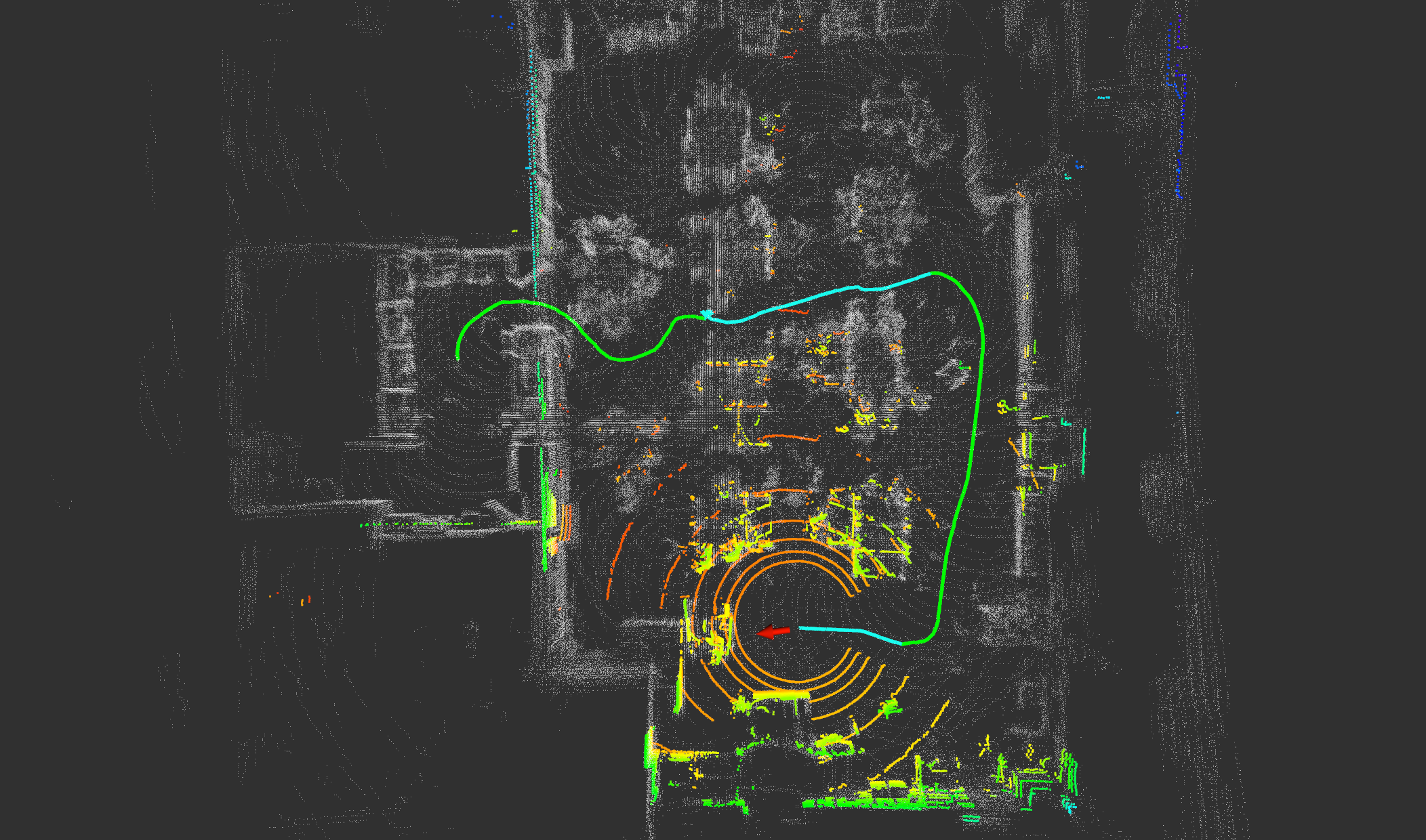}
}%
\subcaption{Second semi-autonomous waypoint}
\end{subfigure}
\caption{Localization with Occlusion in Stevens' ABS Engineering Center. The Red Arrow indicates autonomous navigation goal. The Path in green was manually driven by remote control, while the path in blue was autonomous. The White point cloud is a global reference map, and the colorful pointcloud is the current Velodyne scan. \label{fig:local}}
\end{figure*}

\begin{figure*}[!btp]
\begin{subfigure}[t]{0.24\textwidth}
\centering{%
  \includegraphics[trim={6cm 0 6cm 0},clip,width=\textwidth]{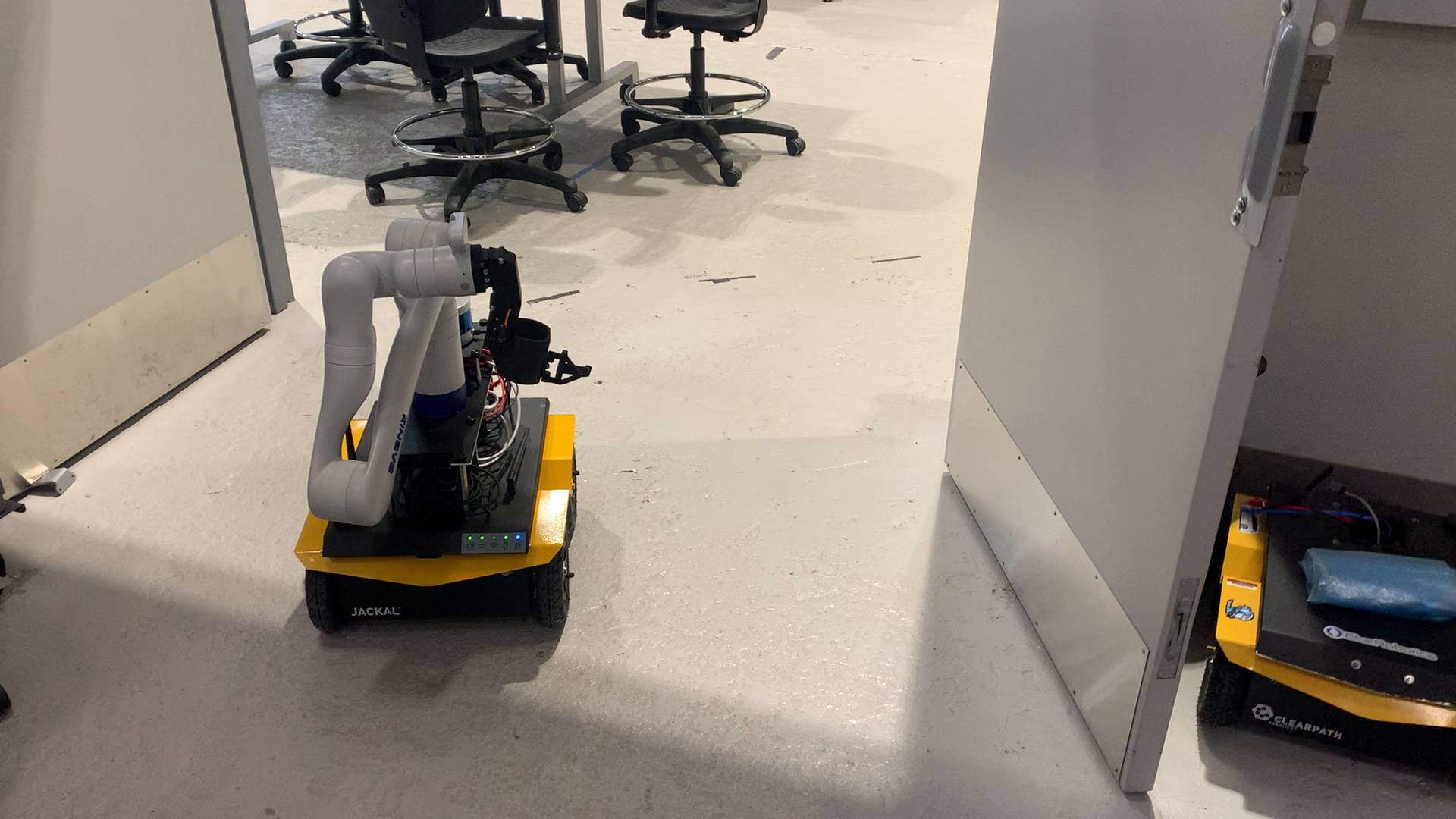}
}%
\subcaption{Compute dist. \& angle to wall\label{fig:a}}
\end{subfigure}%
\hfill
\begin{subfigure}[t]{0.24\textwidth}
\centering{%
\includegraphics[trim={6cm 0 6cm 0},clip,width=\textwidth]{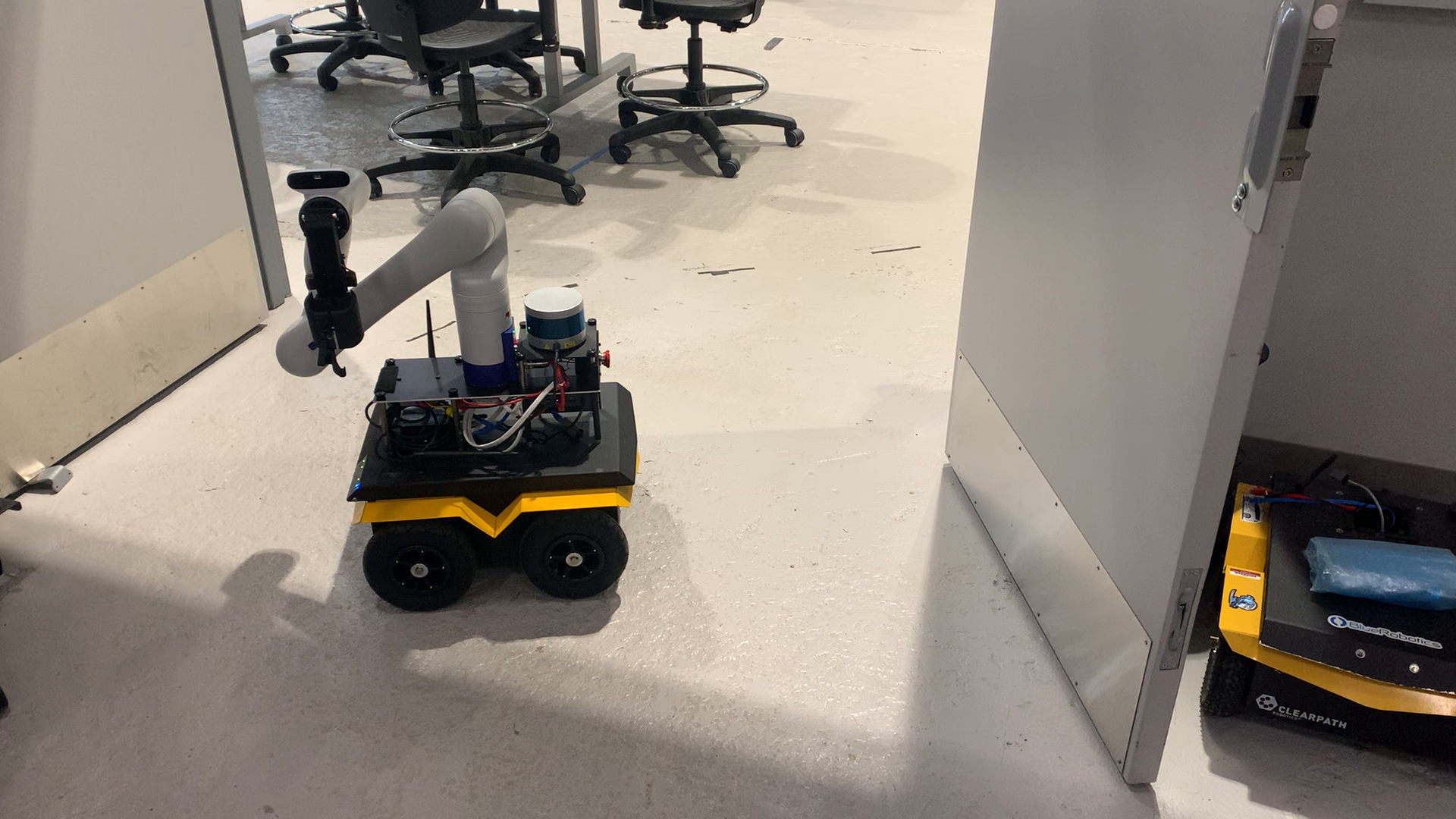}
}%
\subcaption{Turn orthogonal to wall\label{fig:b}}
\end{subfigure}
\hfill
\begin{subfigure}[t]{0.24\textwidth}
\centering{%
\includegraphics[trim={2cm 0 10cm 0},clip,width=\textwidth]{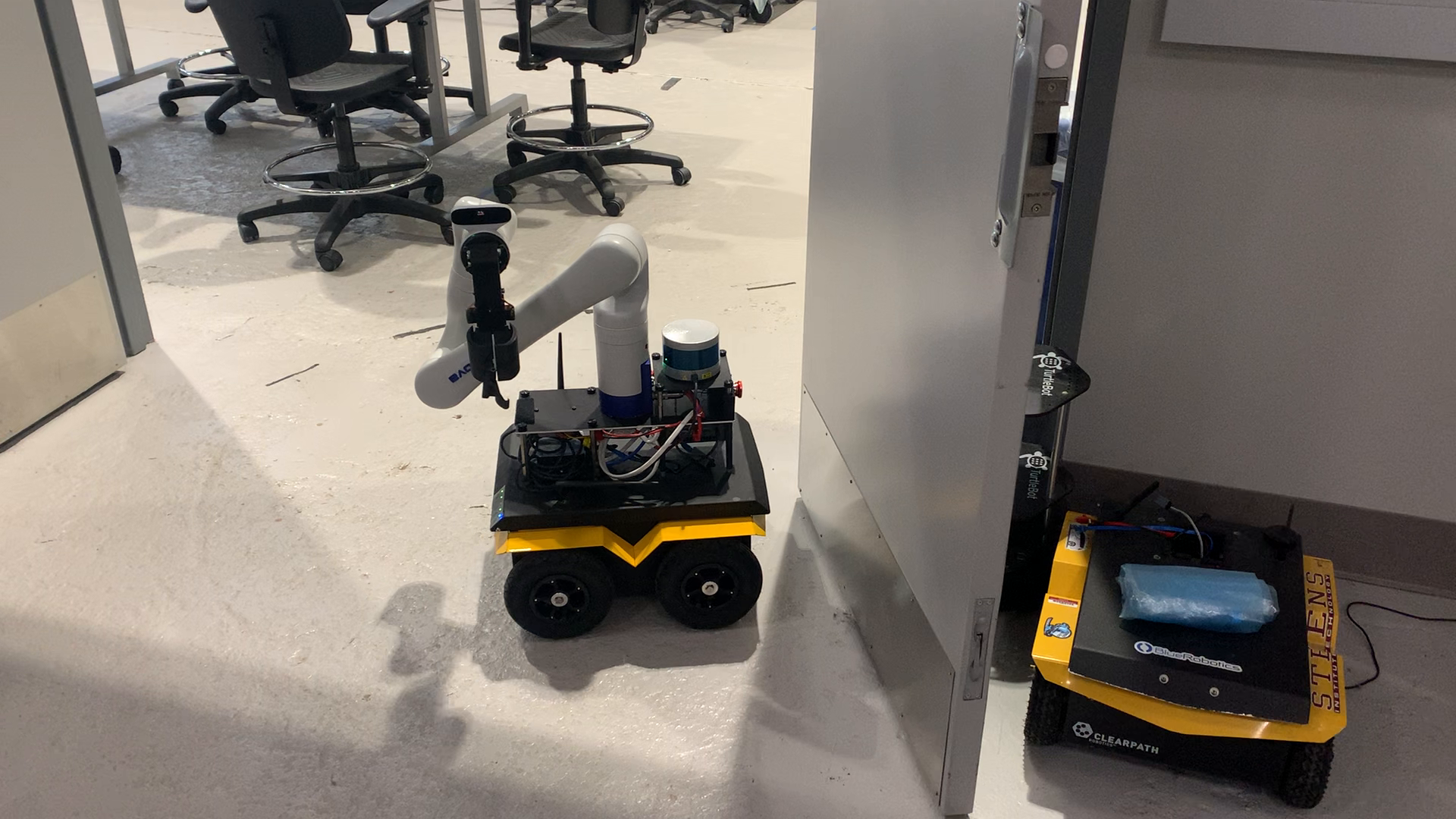}}%
\subcaption{Drive to wall using localization\label{fig:c}}
\end{subfigure}
\hfill
\begin{subfigure}[t]{0.24\textwidth}
\centering{%
\includegraphics[trim={0 0 12cm 0},clip,width=\textwidth]{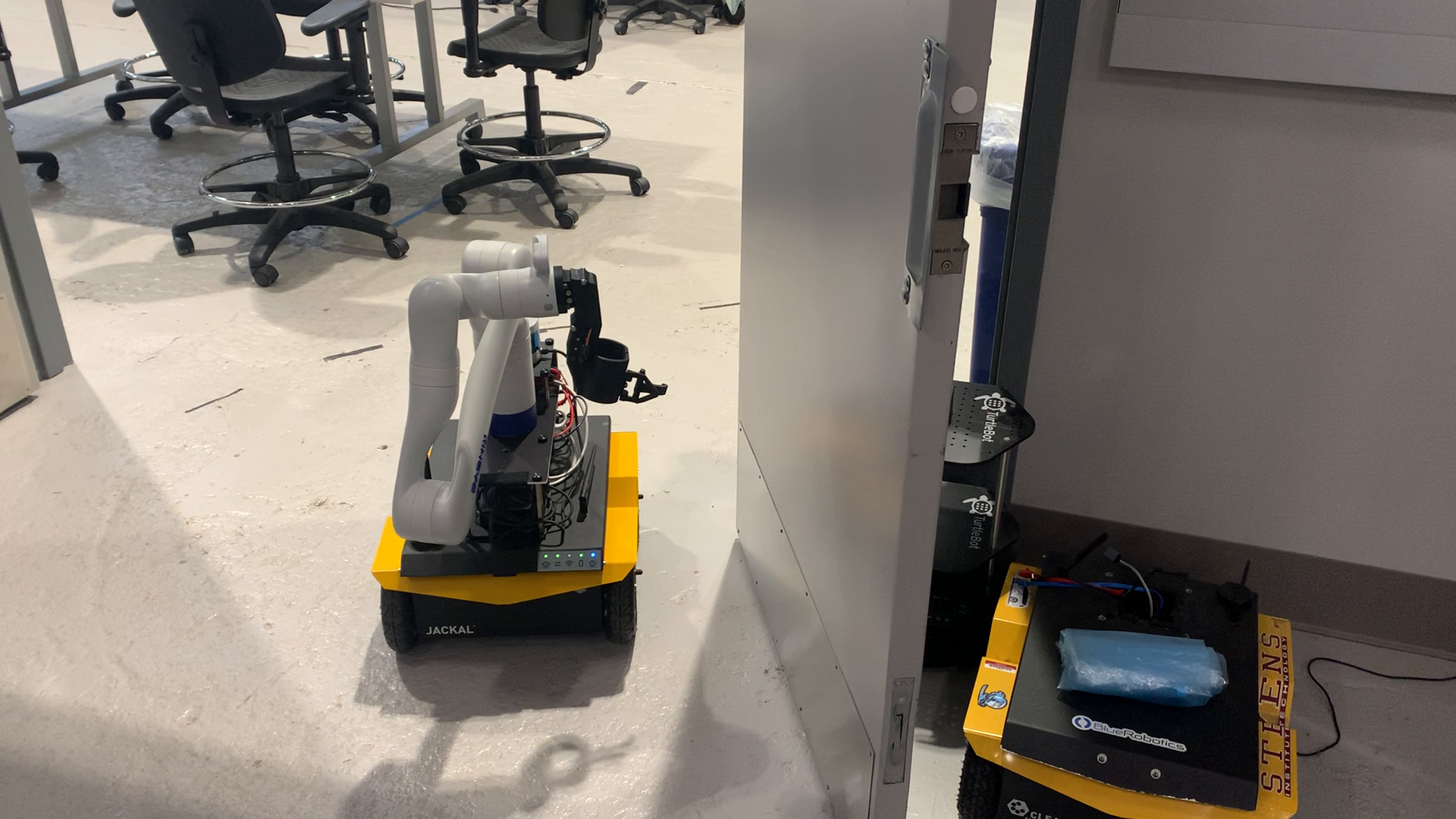}}%
\subcaption{Turn parallel to wall\label{fig:d}}
\end{subfigure}
\caption{Autonomous alignment with a panel through combined action among the Jackal and arm. \label{fig:align}}
\end{figure*}

For safety, a few software considerations were implemented. The most consequential is the ``driving configuration" of the Gen3 manipulator. Seen in Fig.~\ref{fig:travel}, the arm is lowered to reduce the center of gravity while aiming the camera to the side. This enables users to have a clear view along the side of the robot for potential interactions while maintaining stability as the Jackal navigates throughout its environment. A secondary software implementation is acceleration limits when the Jackal drives around. Using smaller acceleration limits, we were able to reduce all bouncing motion resulting from the top-heavy Gen3 manipulator. When the Jackal is stationary and the arm is moving, it also has acceleration limits and boundary limits. While the $17kg$ UGV has a wide enough base to handle the full reach of the $7kg$ manipulator without tipping, extra weight mounted to the wrist could become a concern. Therefore, a software limit can be imposed upon the manipulator workspace to ensure the arm remains stable.
The combination of these 
settings enables the mobile manipulator to freely traverse a cluttered environment with minimal vehicle footprint, while maintaining a large reach for manipulation tasks.

\section{Experimental Results and Discussion}

\subsection*{Localization with Occlusion}

Two main tests were run with this hardware. The first experiment tested the platform's localization capability using the Velodyne puck LiDAR with the occlusion from the Gen3 manipulator included. Most localization performed with a LiDAR will not have such a large portion of its field of view missing. However, even with the occlusion, localization was performed accurately and reliably using our autonomous waypoint navigation package \cite{pearson2023robust}\footnote{https://github.com/RobustFieldAutonomyLab/waypoint\_navigation}. Localization with occlusion was tested in Stevens' ABS Engineering Center and can be seen in Fig.~\ref{fig:local}. 

This environment was a very busy indoor atrium with upwards of 40 students working and moving throughout testing. The white point cloud data in both images was used as the global reference map for localization, and the data used to create that map was gathered approximately an hour beforehand. While testing the localization, both manual driving and autonomous driving of the mobile manipulation platform occurred. Due to the busy nature of dynamic obstacles in this environment, the autonomous driving portion occurred with fewer obstacles nearby.

Mounting the Gen3 manipulator onto the Jackal and driving around in the travel configuration resulted in a Velodyne occlusion of approximately 60 degrees, or one sixth of the complete field of view. To minimize potential failures, the occlusion was engineered to be towards the rear of the vehicle, ensuring obstacle avoidance is possible when necessary. With the occlusion limitation, localization was successfully tested with both manual driving and autonomous waypoint navigation in a busy indoor environment.

\subsection*{Panel Alignment}

The second experiment tested the connection between the two robotic subsystems. For future manipulation tasks, aligning the Jackal and Gen3 arm to a wall panel will be required. While it may be possible to perform this task through the Jackal's Velodyne localization alone, the accuracy and precision may not be reliable enough for all scenarios. Therefore, using a combination of the arm's wrist-mounted RGB-D camera and the Jackal's Velodyne localization can produce the desired effect.

This behavior was tested on a propped open door, to show the effectiveness of alignment on partial planar structures. We assume the camera is facing some panel-like object for this process to succeed. The first step uses the depth camera data to identify both the distance and the angle between the current position of the robots and the identified panel. Once computed, the Jackal rotates towards the panel until orthogonal, using its localization to the global reference frame to ensure accurate orientation. The Jackal then proceeds to drive towards the panel, and its standoff distance is also checked using localization in the global reference frame to ensure an accurate traversal. The final step for panel alignment is for the Jackal to return to being approximately parallel to the panel to enable the maximum workspace for the Kinova manipulator to begin working as intended. This also allows the program to confirm that the robots are close enough to begin the manipulation task (if they are not, a final rotation is performed by the arm to further align with the panel). This process is summarized in Fig. \ref{fig:align}.  We expect this functionality to be used for placing the Jackal at specific locations in an indoor environment where manipulation tasks are required. The Gen3 arm should complete all tasks while the Jackal is stationary, before returning to the travel configuration and moving to the next location.


\section{Conclusion}
A custom combination of Clearpath's Jackal UGV and Kinova's Gen3 six degree-of-freedom manipulator has been proposed for indoor inspection tasks in confined areas requiring a larger manipulator reach. With a safe travel configuration and limited accelerations, the platform was stable during autonomous navigation throughout a variety of indoor environments. The hardwired data connection between robots ensures no data loss when running computationally intensive programs for solving navigation or identification tasks. Further studies with this platform will include utilization of the image data from the RGB-D wrist mounted camera, such as detection, tracking, and visual servoing tasks. Custom end effectors will define the set of tasks that can be accomplished by this platform, which hopefully will make use of its autonomous navigation and long reach capabilities.


\section*{Acknowledgments}
This work was supported by a grant from the Consolidated Edison Company of New York, Inc.



\bibliographystyle{asmeconf}  
\bibliography{bibliography}


\end{document}